\documentclass{article}
\usepackage{ijcai24}

\usepackage{times}
\usepackage{soul}
\usepackage{url}
\usepackage[hidelinks]{hyperref}
\usepackage[utf8]{inputenc}
\usepackage[small]{caption}
\usepackage{graphicx}
\usepackage{amsmath}
\usepackage{amsthm}
\usepackage{booktabs}
\usepackage{algorithm}
\usepackage{algorithmic}
\usepackage[switch]{lineno}

\usepackage{subfig}
\usepackage{multirow}
\usepackage{adjustbox}

\usepackage{colortbl}
\usepackage[table]{xcolor}
\usepackage{soul}
\sethlcolor{cyan!20}

\title{Employing Layerwised Unsupervised Learning to Lessen Data and Loss Requirements in Forward-Forward Algorithms}

\author{
Taewook Hwang$^1$
\and
Hyein Seo$^2$\and
Sangkeun Jung$^3$\\
\affiliations
$^{1,2,3}$Department of Computer Science and Engineering, Chungnam National University\\
\emails
\{taewook5295,hyenee97\}@gmail.com,
hugman@cnu.ac.kr
}

\begin{document}

\maketitle

\begin{abstract}

Recent deep learning models such as ChatGPT utilizing the back-propagation algorithm have exhibited remarkable performance. However, the disparity between the biological brain processes and the back-propagation algorithm has been noted. The Forward-Forward algorithm, which trains deep learning models solely through the forward pass, has emerged to address this. Although the Forward-Forward algorithm cannot replace back-propagation due to limitations such as having to use special input and loss functions, it has the potential to be useful in special situations where back-propagation is difficult to use. To work around this limitation and verify usability, we propose an Unsupervised Forward-Forward algorithm. Using an unsupervised learning model enables training with usual loss functions and inputs without restriction. Through this approach, we lead to stable learning and enable versatile utilization across various datasets and tasks. From a usability perspective, given the characteristics of the Forward-Forward algorithm and the advantages of the proposed method, we anticipate its practical application even in scenarios such as federated learning, where deep learning layers need to be trained separately in physically distributed environments.

\end{abstract}


\section{Introduction} \label{sec:Introduction}

Recently, large language models, including ChatGPT, have demonstrated performance surpassing that of human capabilities. This achievement is largely attributed to the back-propagation algorithm(BP) \cite{bp60,bp70}, a highly effective learning algorithm, complemented by advancements in hardware technology. However, one of the criticisms of BP lies in their deviation from the biological processes underlying brain behavior \cite{bpcritic87,bpcritic89,bpcritic90,bpcritic16}. Also, because BP utilizes a backward pass, it has the limitation that all the layers that make up the model must be connected in real time during the training process.

Despite numerous learning algorithms being proposed to address these issues, none have successfully supplanted BP due to performance and versatility constraints \cite{pc,fa,ff}. However, recently, \cite{ff} proposed a new learning algorithm named the Forward-Forward algorithm(FF) as a key figure in the development of BP. In contrast of BP, the FF addresses this issue by employing two forward passes, eliminating the need for a backward pass.

However, FF also presents its own limitations. \textit{The FF diverges from traditional deep learning methodologies by requiring three inputs and a specialized loss function.} Although these constraints prevent the FF from outright replacing BP, it can be a potential alternative in specific scenarios where traditional BP may not be optimal.

\begin{figure*}[h]
\centering
\subfloat[Original data used for the proposed Unsupervised Forward-Forward algorithm and the Positive, Negative, and Neutral data used for the Forward-Forward algorithm.]{\label{subfig:data}\includegraphics[scale=.5]{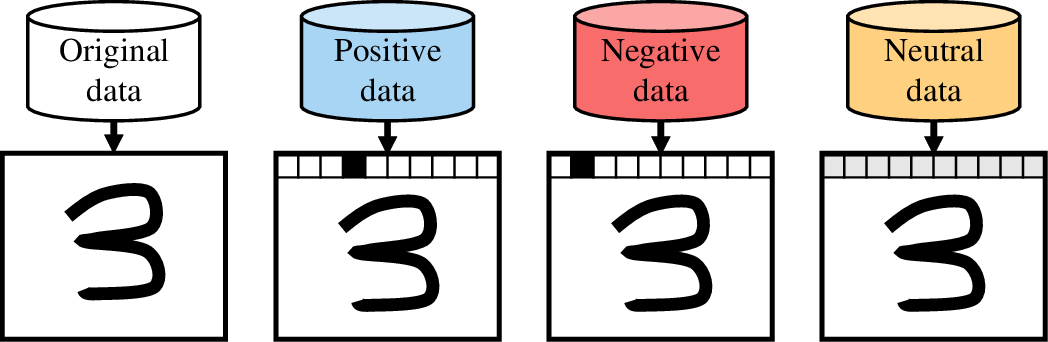}}

\begin{minipage}{.5\linewidth}
\centering
\subfloat[Overview of Forward-Forward algorithm]{\label{subfig:ff}\includegraphics[scale=.5]{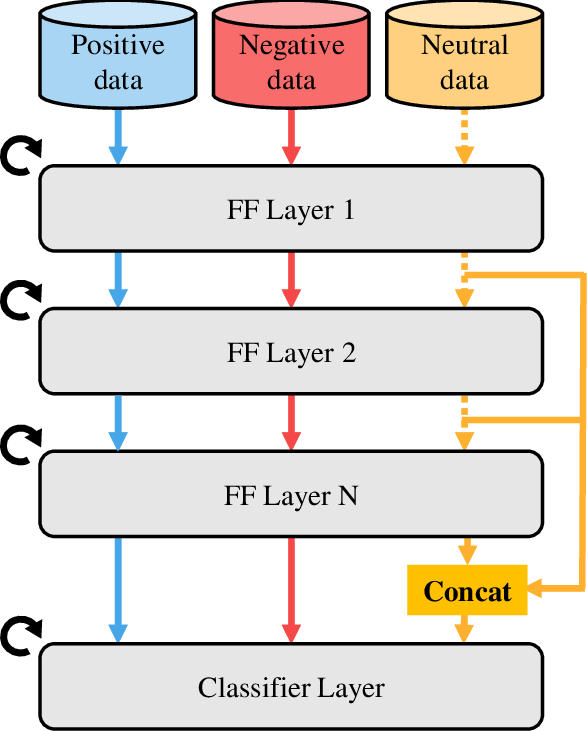}}
\end{minipage}%
\begin{minipage}{.5\linewidth}
\centering
\subfloat[Overview of Unsupervised Forward-Forward algorithm]{\label{subfig:uff}\includegraphics[scale=.5]{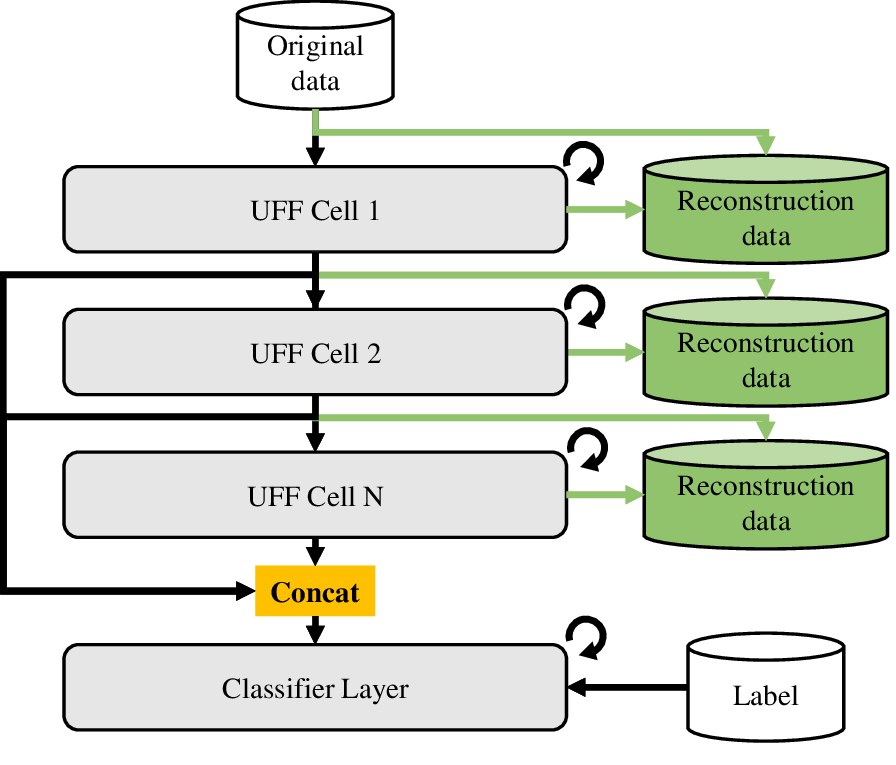}}
\end{minipage}\par\medskip

\caption{Using Data and Forward-Forward algorithm and Unsupervised learning Forward-Forward model.}
\label{fig:model}
\end{figure*}

To address the limitation of FF and validate features of FF, \textit{we propose the adoption of a compact unsupervised learning model as a substitute for the individual layers typically used in the FF.} In this paper, the proposed model is referred to as the \textbf{Unsupervised learning Forward-Forward model(UFF)}. This approach, leveraging an unsupervised learning model trainable through input reconstruction, eliminates the need for the specialized inputs and loss function used in the FF. Consequently, the inputs can be the same as those used in standard deep learning process.

Additionally, a distinctive characteristic of the FF is the local and independent training of each layer. This feature that permits physical separation of layers during both learning and inference phases, while still maintaining the capability for inter-layer information exchange. In our proposed methodology, we focus on confirming that when a using unsupervised model, is employed instead of a single layer, termed `cell' in this paper, as in the original FF, it is adept at information transfer between each of these cells.




The main contributions of our paper are as follows:

\begin{itemize}
    \item Overcoming the limitations inherent in the Forward-Forward algorithm through using an unsupervised learning model.
    \item Searching unsupervised learning models that exhibit stable learning in the proposed method.
    \item Validating the scalability of employing a composite of layers or models, instead of a singular layer.
\end{itemize}

Consequently, compared to traditional models employing BP, the UFF approach is expected to provide greater flexibility in adding, deleting, or replacing layers post-training. In addition, it is expected to be highly useful in environments where layers must be trained in physical separation, such as federated learning.


Figure \ref{fig:model} presents a comprehensive overview. Figure \ref{subfig:data} illustrates the input data utilized by the FF and UFF. Figure \ref{subfig:ff} provides a comprehensive overview of the FF itself. Figure \ref{subfig:uff} depicts an overview of the UFF model, as proposed in this study. UFF cells consist of independent unsupervised learning models, such as AutoEncoders(AE) \cite{autoencoder} or Generative Adversarial Networks(GAN) \cite{gan}. The hidden vector from each cell serves as the input for the subsequent cell. This structure facilitates the independent identification of crucial information utilized by each cell.

\begin{figure}[ht!]
\centering
\subfloat[Auto-Encoder cell]{\label{subfig:aeff}
\includegraphics[scale=.7]{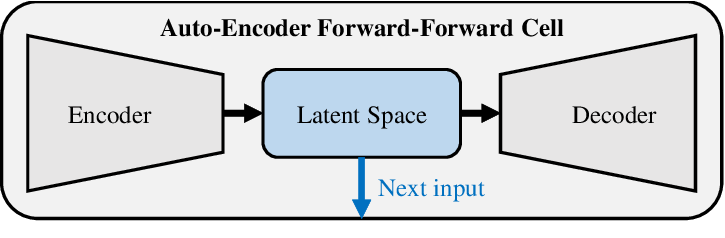}}

\subfloat[Denoising Auto-Encoder cell]{\label{subfig:daeff}
\includegraphics[scale=.7]{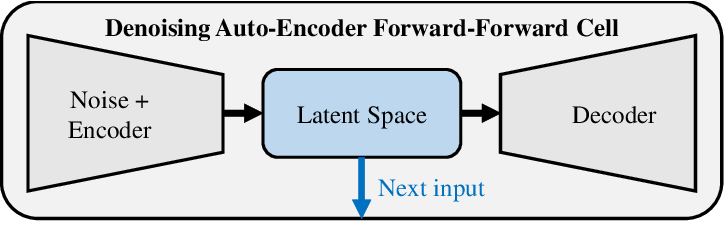}}

\subfloat[Convolutional Auto-Encoder cell]{\label{subfig:caeff}
\includegraphics[scale=.7]{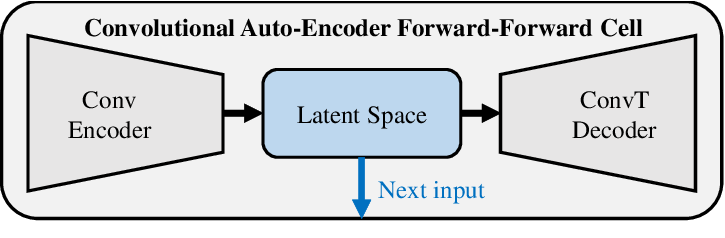}}

\subfloat[Generative Adversarial Network cell]{\label{subfig:ganff}
\includegraphics[scale=.7]{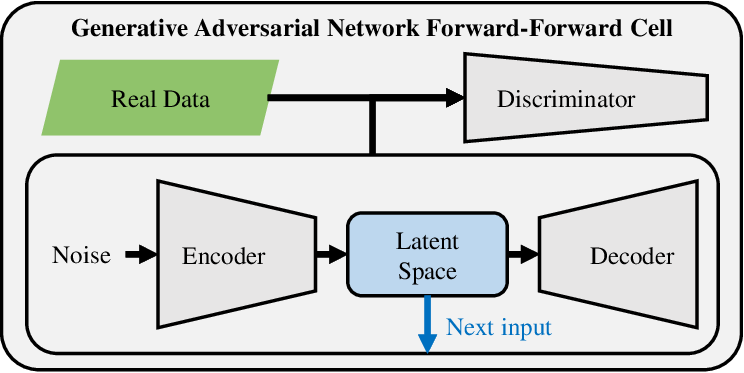}}
\caption{Unsupervised deep learning model used in Unsupervised Forward-Forward cell.}
\label{fig:layer}
\end{figure}

\section{Related Works} \label{sec:RelatedWorks}

The criticisms raised against BP from a biological perspective are primarily as follows. In biological neural networks,  the direction of neuron stimulus transmission is unidirectional, making a backward pass challenging. Although reciprocal connections \cite{rc62,rc94} display elements of backward transmission, they are restricted to local connections between adjacent neurons, complicating the development of global feedback pathways as seen in BP. Additionally, the weights in the forward pass synapses are not symmetrical with those in the backward pass, complicating the execution of feedback. To address these limitations, Feedback Alignment \cite{fa} and Predictive Coding \cite{pc} have emerged.

Feedback Alignment is a method that utilizes random weights for feedback during the weight update process.This approach demonstrated the potential for learning in situations where the weights in the synapses forward pass and backward pass are not symmetrical. BP involves complex calculations to synchronize the weights of the forward and backward passes for updates. However, Feedback Alignment suggests that comparable learning to BP can be achieved with random values, provided that the directionality of the weight updates remains consistent. This directionality is refined through iterative learning, allowing deep learning models to progressively identify appropriate values. Therefore, it can be posited that BP, diverging from a biological perspective, imposes excessively strict updates, and Feedback Alignment mitigates some of the biological criticisms directed at BP.

Predictive Coding is a method that implements local feedback, where a current layer predicts the input for the next layer and then calculates the loss by comparing the actual input with the predicted value. This approach enables weight updates between adjacent layers. \cite{pc17,pc22} have shown that the Predictive Coding method achieves performance similar to BP in various models such as Multi Layer Perceptrons(MLPs) \cite{mlp}, Convolutional Neural Networks(CNN) \cite{cnn}, and Recurrent Neural Networks \cite{rnn}.

Following the release of FF, \cite{pff} is research that combines Predictive Coding with FF. The FF algorithm uses only the forward pass, making it similar to the brain's operational method. However, it requires verification to ensure effective information transfer between adjacent layers. \cite{pff} employed Predictive Coding to facilitate local learning among layers, demonstrating performance comparable to MLPs using BP.


\section{Unsupervised Forward-Forward Algorithm} \label{sec:UFFM}

We propose the UFF learning approach, utilizing unsupervised deep learning models. Our method eliminates the constraints on inputs and loss calculations present in FF, thereby allowing for the adoption of input formats and loss computations typical of general deep learning models. Through this algorithm, our goal is to maintain the learning direction of FF while seeking compatibility with existing deep learning models, resulting in a versatile learning approach. We anticipate that our proposed method will demonstrate high utility in scenarios where physical layer separation makes the use of BP challenging.

\subsection{Architecture} \label{subsec:ModelArchitecture}

The overall structure of the model is depicted in Figure \ref{fig:model}. Original data is input into the first cell, and each cell engages in unsupervised learning, similar to Auto-Encoders and GANs, by attempting to reconstruct the input. This architecture is flexible and can be applied to any system where local loss can be calculated independently at each cell, regardless of whether the input is reconstructed or not, provided that it produces latent vectors of a uniform size.

The unsupervised learning models utilized in this study include Auto-Encoders(AE), Denoising Auto-Encoders(DAE) \cite{dae}, Convolutional Auto-Encoders(CAE), and Generative Adversarial Networks(GANs). Figure \ref{fig:layer} presents the architecture where each unsupervised learning model comprises cells. Figure \ref{subfig:aeff} is a cell using AE, and Figure \ref{subfig:daeff} is a model using AE with some random noise added to the input. Figure \ref{subfig:caeff} shows an AE model composed of a convolutional encoder and a transposed convolutional decoder. Figure \ref{subfig:ganff} is a cell that utilizes the structure of GAN, where the generator is composed of the structure of AE, and the discriminator compresses the fake data and real data generated by the generator from random noise into one-dimensional vectors to determine the authenticity.

Each unsupervised learning model is implemented using the minimum number of deep learning layers. We fixed the size of the latent vector at half the size of the input to ensure computational convenience. Although there is no limit on the number of cells, if additional cells are added when the latent vector size cannot be reduced further, we maintained the latent vector size of the added cell equal to the input size.

The last classifier layer serves as the layer that outputs the overall results of the model. It is a general fully-connected layer not an unsupervised learning model and can be composed of different types of layers depending on the task. The last layer employs concatenation of the latent vectors from all cells as its input.

\subsection{Model Training} \label{subsec:ModelLearning}

The cells composed of AE series models function such that the encoder transforms input data into a latent vector, and the decoder is trained to generate data from this latent vector in a way that it becomes identical to the input. During this process, the latent vector generated by the encoder is utilized as the input for the subsequent cell.

The cells composed of GANs use random noise as input for their generators. Each encoder within the generators generates a latent vector from this input, and then the decoder of the generators produces fake data from these latent vectors. The discriminator receives both real and fake data, learning to distinguish which is real. Following this, the generators create new fake data from random noise, which the discriminators then assess, using the calculated loss for further training. After this process, the latent vector generated by the encoders of the generators, using inputs from the previous cell, is passed on as the input for the subsequent cell.

The last layer of each model is trained to predict the answers. The training approach varies based on the task, including classification, regression, and generation. Furthermore, this layer is also locally trained.

The loss function for the task of input reconstruction employed mean squared error. For the discriminator in GANs, we used binary cross-entropy loss. The classifier layer utilized cross-entropy loss. For all models, AdamW \cite{adamw} was selected as the optimizer, and the ReLU \cite{relu} was employed as the activation function.

Layer normalization \cite{layernorm} is applied to every input of each layer to normalize the data distribution. This was done to mitigate gradient vanishing and exploding issue, and achieve more stable and faster learning and improved generalization.


\begin{table*}[]
\centering
\begin{tabular}{cc|ccc|c|cccc}
\hline
\multicolumn{2}{c|}{\textbf{MNIST}}     & \textbf{SLP}    & \textbf{MLP}    & \textbf{CNN}             & \textbf{FF}     & \textbf{AEFF}   & \textbf{DAEFF}  & \textbf{CAEFF}           & \textbf{GANFF}  \\ \hline
\multirow{4}{*}{\textbf{Sequence training}}   & 2 layers & \multirow{4}{*}{0.9274} & \textbf{0.9875} & 0.9921                   & 0.9782          & 0.9689          & 0.9723          & 0.9791                   & 0.9569          \\
                                  & 3 layers &                         & 0.9856          & 0.9942                   & \textit{\textbf{0.9795}} & 0.9706          & 0.9736          & \cellcolor{cyan!20}\textit{\textbf{0.9821}} & \textbf{0.966}  \\
                                  & 4 layers &                         & \textbf{0.9875} & 0.9944                   & 0.9751          & 0.9704          & 0.9722          & 0.9808                   & 0.9581          \\
                                  & 5 layers &                         & 0.987           & \textit{\textbf{0.9951}} & 0.9789          & \textbf{0.9713} & \textbf{0.9737} & 0.9793                   & 0.9586          \\ \hline
\multirow{4}{*}{\textbf{Separate training}}   & 2 layers & \multirow{4}{*}{0.9274} & \textbf{0.9875} & 0.9921                   & 0.9687          & 0.9707          & 0.9679          & \cellcolor{cyan!20}\textit{\textbf{0.9828}} & \textbf{0.9621} \\
                                  & 3 layers &                         & 0.9856          & 0.9942                   & 0.9674          & 0.9704          & 0.9710          & 0.9813                   & 0.9604          \\
                                  & 4 layers &                         & \textbf{0.9875} & 0.9944                   & 0.9694          & \textbf{0.9731} & 0.9721          & 0.9801                   & 0.9598          \\
                                  & 5 layers &                         & 0.9870          & \textit{\textbf{0.9951}} & \textit{\textbf{0.9700}} & 0.9717          & \textbf{0.9729} & 0.9824                   & 0.9598          \\ \hline
\end{tabular}
\caption{Best performances on the MNIST dataset. \textbf{Bold} is best of each model, \textit{\textbf{italic}} is best of each learning method, and \hl{\textbf{\textit{highlight}}} is the best of our proposed models.} \label{tab:best_mnist}
\end{table*}

\begin{table*}[]
\centering
\begin{tabular}{cc|ccc|c|cccc}
\hline
\multicolumn{2}{c|}{\textbf{CIFAR10}}     & \textbf{SLP}    & \textbf{MLP}    & \textbf{CNN}             & \textbf{FF}     & \textbf{AEFF}   & \textbf{DAEFF}  & \textbf{CAEFF}           & \textbf{GANFF}  \\ \hline
\multirow{4}{*}{\textbf{Sequence training}} & 2 layers & \multirow{4}{*}{0.3895} & \textbf{0.5439} & 0.6979                   & \textit{\textbf{0.4886}} & 0.4705          & \textbf{0.4691} & 0.5704                   & 0.4467          \\
                                  & 3 layers &                         & 0.5393          & 0.7645                   & 0.4713          & \textbf{0.4751} & 0.4474          & \cellcolor{cyan!20}\textit{\textbf{0.5726}} & 0.4394          \\
                                  & 4 layers &                         & 0.5364          & 0.7598                   & 0.4751          & 0.4745          & 0.4573          & 0.5219                   & \textbf{0.4542} \\
                                  & 5 layers &                         & 0.5205          & \textit{\textbf{0.777}}  & 0.462           & 0.473           & 0.4535          & 0.5394                   & 0.4436          \\ \hline
\multirow{4}{*}{\textbf{Separate training}} & 2 layers & \multirow{4}{*}{0.3895} & \textbf{0.5439} & 0.6979                   & 0.3308          & 0.4658          & 0.4539          & 0.5670                   & 0.4420          \\
                                  & 3 layers &                         & 0.5393          & 0.7645                   & 0.3508          & 0.4771          & 0.3788          & 0.5753                   & 0.4410          \\
                                  & 4 layers &                         & 0.5364          & 0.7598                   & 0.3939          & \textbf{0.4817} & \textbf{0.4772} & \cellcolor{cyan!20}\textit{\textbf{0.6057}} & \textbf{0.4573} \\
                                  & 5 layers &                         & 0.5205          & \textit{\textbf{0.777}}  & \textit{\textbf{0.3987}} & 0.4813          & 0.4621          & 0.5705                   & 0.4367          \\ \hline
\end{tabular}
\caption{Best performances on the CIFAR10 dataset. \textbf{Bold} is best of each model, \textit{\textbf{italic}} is best of each learning method, and \hl{\textbf{\textit{highlight}}} is the best of our proposed models.} \label{tab:best_cifar}
\end{table*}

\section{Experiments} \label{sec:Experiments}

Our goal is to validate the experiment that proposed UFF models can replace traditional FF models in terms of performance, and to evaluate their compatibility with various general unsupervised learning methods. Additionally, we aim to confirm whether these models can demonstrate stable and consistent performance under conditions similar to BP. Finally, Our goal includes investigating the extent of performance variation depending on the number of cells and quantifying the difference in performance between these models and a Single Layer Perceptron(SLP).

\subsection{Models} \label{subsec:Models}


We employed baselines consisting of SLP, MLP, and CNN models trained using BP, along with FF models that we reproduced. Furthermore, we designed our experiments using four UFF models as layers: Auto-Encoder Forward-Forward(AEFF), Denoising Auto-Encoder Forward-Forward(DAEFF), Convolutional Auto-Encoder Forward-Forward(CAEFF), and Generative Adversarial Network Forward-Forward(GANFF). Throughout the experimental process, we maintained consistency in the configuration of optimizers, hidden dimensions, and the number of cells used.


For SLP, a single fully-connected layer was employed to generate outputs directly from the input data. In the MLP and FF models, each layer consisted of one fully-connected layer and a ReLU activation function. The AEFF, DAEFF, and GANFF models utilized a one fully-connected layer and ReLU in each encoder and decoder layer. The discriminator in GANFF was designed with a fully-connected layer and ReLU to convert the input into a hidden vector, and another fully-connected layer to transform the hidden vector into an output.

In the CNN and CAEFF, each layer included a convolution, ReLU, and max pooling. For convolutional layers, a kernel size of 3, stride of 1, and padding of 1 were used, and max pooling had a kernel size of 2. The decoder of CAEFF employed transpose convolution for reconstruction, with a kernel size of 2 and stride of 2. When using the MNIST dataset, which has a width and height of 28, passing through two CAEFF layers results in widths and heights of 7, an odd number. Under these circumstances, a convolution kernel size of 2 was used, with the transpose convolution having a kernel size of 4 and a stride of 1.

\subsection{Datasets} \label{subsec:Datasets}

The datasets employed in our experiment were MNIST and CIFAR10. MNIST comprises a total of 10 labels, with 60,000 training data samples and 10,000 test data samples. Similarly, CIFAR10 also has 10 labels, consisting of 50,000 training data samples and 10,000 test data samples. In this study, validation data was not utilized.

\subsection{Computing infrastructure} \label{subsec:Computinginfrastructure}

All deep learning models used in this experiment were written in PyTorch, the hardware and software specifications used in this experiment are as follows:

\begin{itemize}
    \item CPU : Intel Xeon Silver 4114 CPU 2.20GHz
    \item GPU : NVIDIA Tesla V100 SXM2 32Gb 
    \item RAM : 512Gb (Only partially utilized)
    \item Framework : PyTorch 2.0.1
\end{itemize}

\subsection{Experimental Setup} \label{subsec:ExperimentalSetup}
In this experiments, we utilized the FF source code\footnote{https://github.com/loeweX/Forward-Forward} implemented in PyTorch.
We utilized the Weights \& Biases(WandB) \cite{wandb} hyper-parameter tuning method known as Sweep to train our models. For each experiment, we conducted 10 runs to measure performance. The batch size was set at 512 for all experiments. For the MNIST dataset, the maximum number of epochs was limited to 100, while for the CIFAR10 dataset, it was set at 200. The noise ratio for the DAEFF was fixed at 0.2. Additionally, the hidden dimension for each model was established at 1,024. In UFF models, the size of the hidden dimension was designed to reduce by half upon each cell pass.


The hyper-parameters explored for the SLP and MLP are as follows:

\begin{itemize}
    \item Learning rate : [1e-3, 1e-5]
    \item Weight decay : [1e-2, 1e-4]
\end{itemize}

The hyper-parameters searched by FF, AEFF, DAEFF, CAEFF, and GANFF are as follows. We set different ranges for the learning rate and weight decay in the final classifier layer compared to the other FF layers and cells. Generally, the FF layers and cells require a sufficiently small learning rate for effective training.

\begin{itemize}
    \item FF layer and cell learning rate : [1e-4, 1e-6]
    \item FF layer and cell weight decay : [1e-3, 1e-5]
    \item Classifier layer learning rate : [1e-3, 1e-5]
    \item Classifier layer weight decay : [1e-2, 1e-4]
\end{itemize}

Additionally, the size of the output channel for the first layer in CNN and CAEFF was configured as follows. It was set to double with each traversal through a convolution layer.

\begin{itemize}
    \item First convolution output channel size : 8, 16, 32, 64
\end{itemize}

The SLP model performed hyper-parameter searching once for each dataset and measured performance for 10 trials. For other models, the number of layers was set to 2, 3, 4, and 5, and hyper-parameter searching was conducted for each layer configuration, resulting in 40 performance measurements across datasets.

When using the FF learning method, each cell is trained independently. Therefore, two training strategies were employed. The first involves sequentially training from the initial cell to the last classifier layer in each epoch, referred to as \textit{sequence training}. The second strategy involves training each cell fully up to the maximum epochs before moving on to the next cell, termed \textit{separate training}. In this experiment, FF, AEFF, DAEFF, CAEFF, and GANFF models were compared in terms of performance using both sequence and separate training.

\begin{figure*}
\centering
\captionsetup{justification=centering}
\subfloat[\label{fig:mnist2} 2 layers]{\includegraphics[width=0.2\paperwidth]{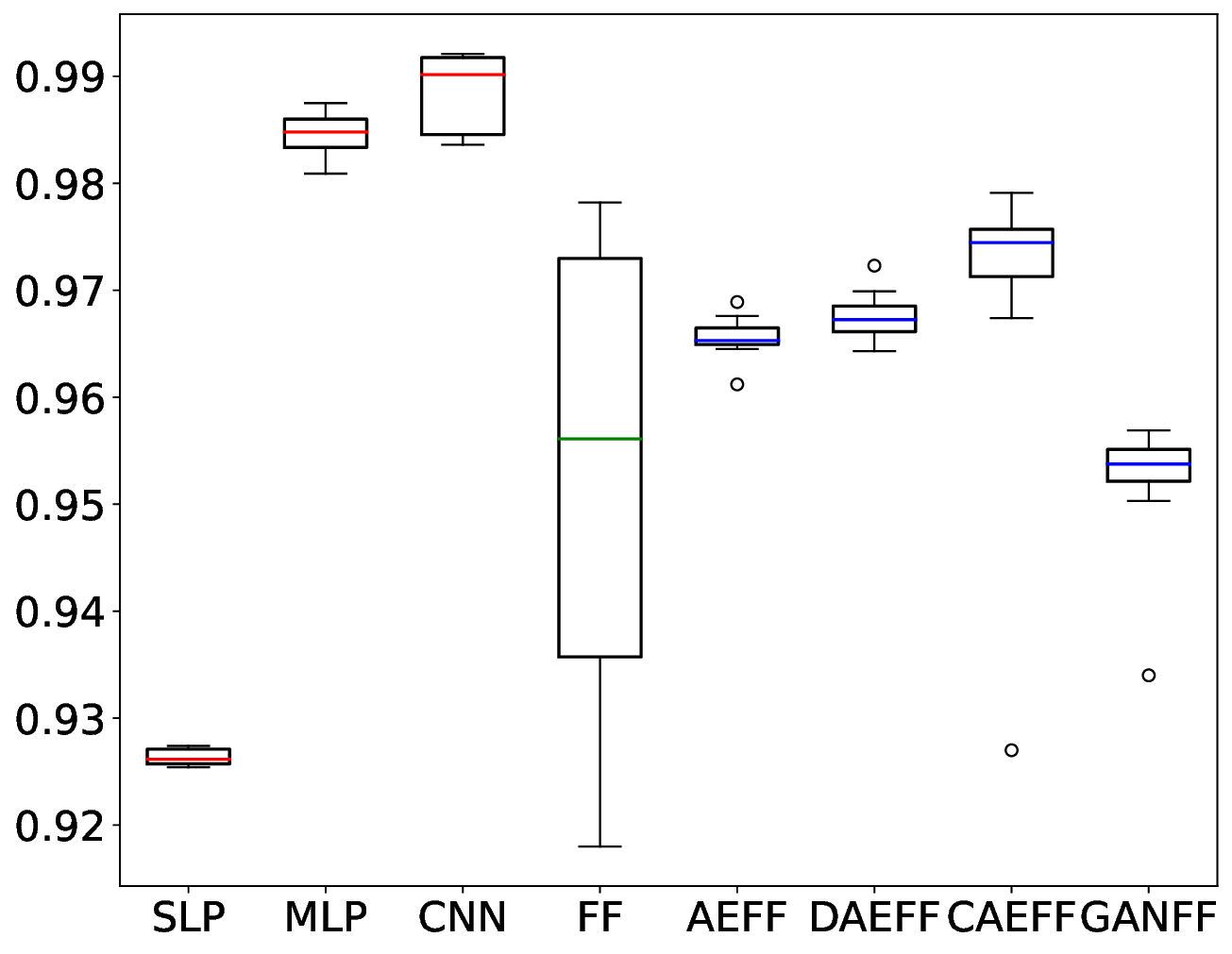}}
\subfloat[\label{fig:mnist3} 3 layers]{\includegraphics[width=0.2\paperwidth]{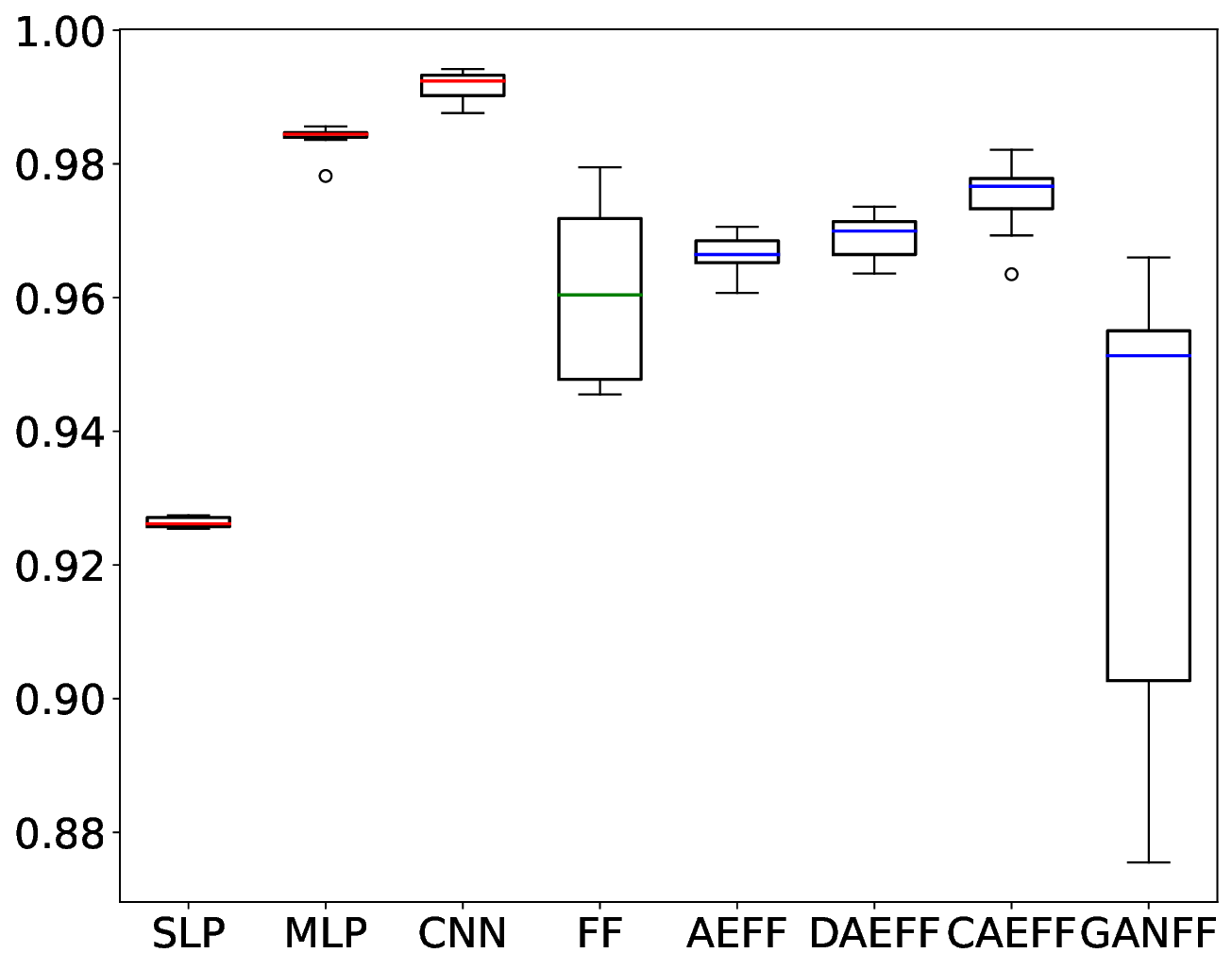}}
\subfloat[\label{fig:mnist4} 4 layers]{\includegraphics[width=0.2\paperwidth]{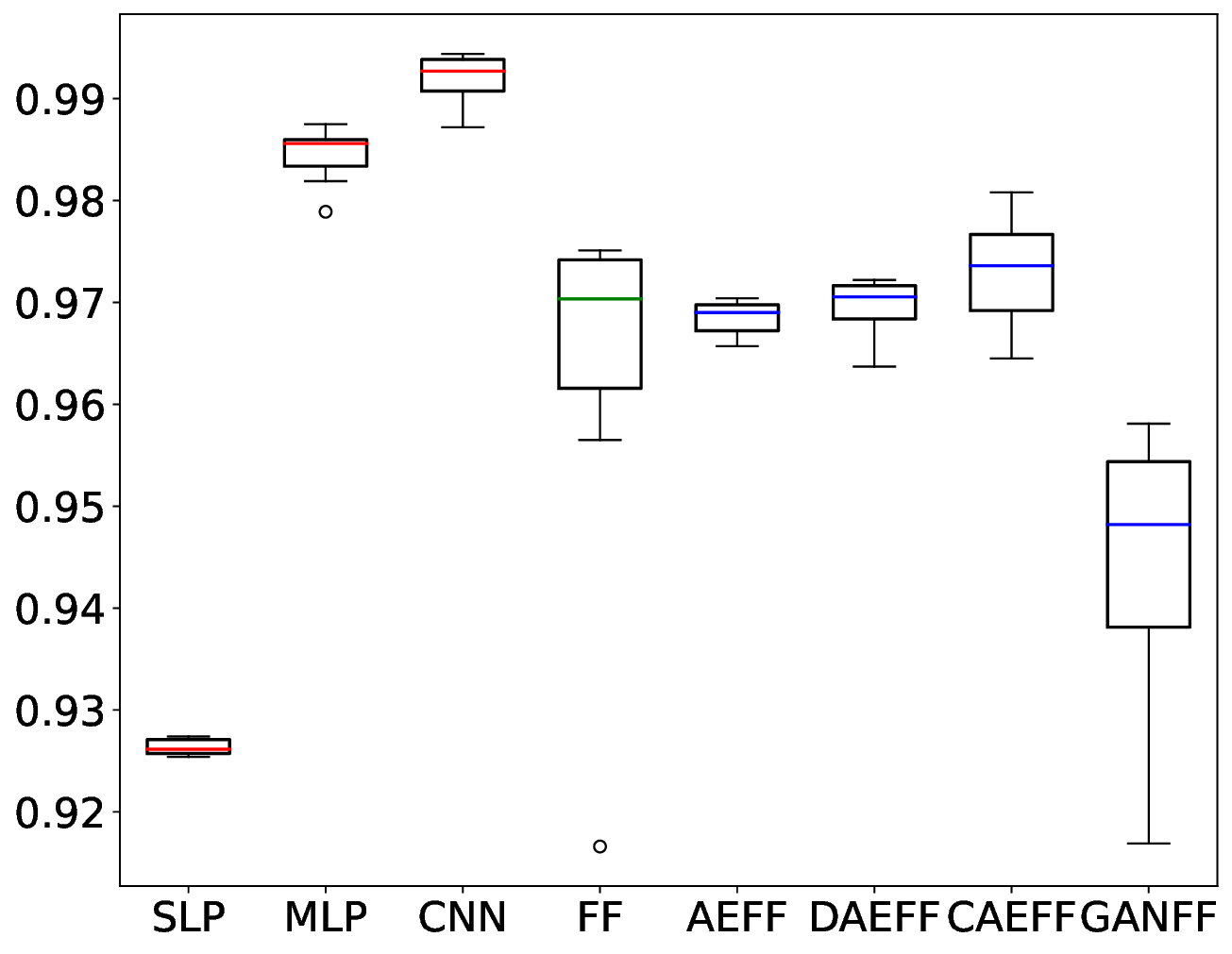}}
\subfloat[\label{fig:mnist5} 5 layers]{\includegraphics[width=0.2\paperwidth]{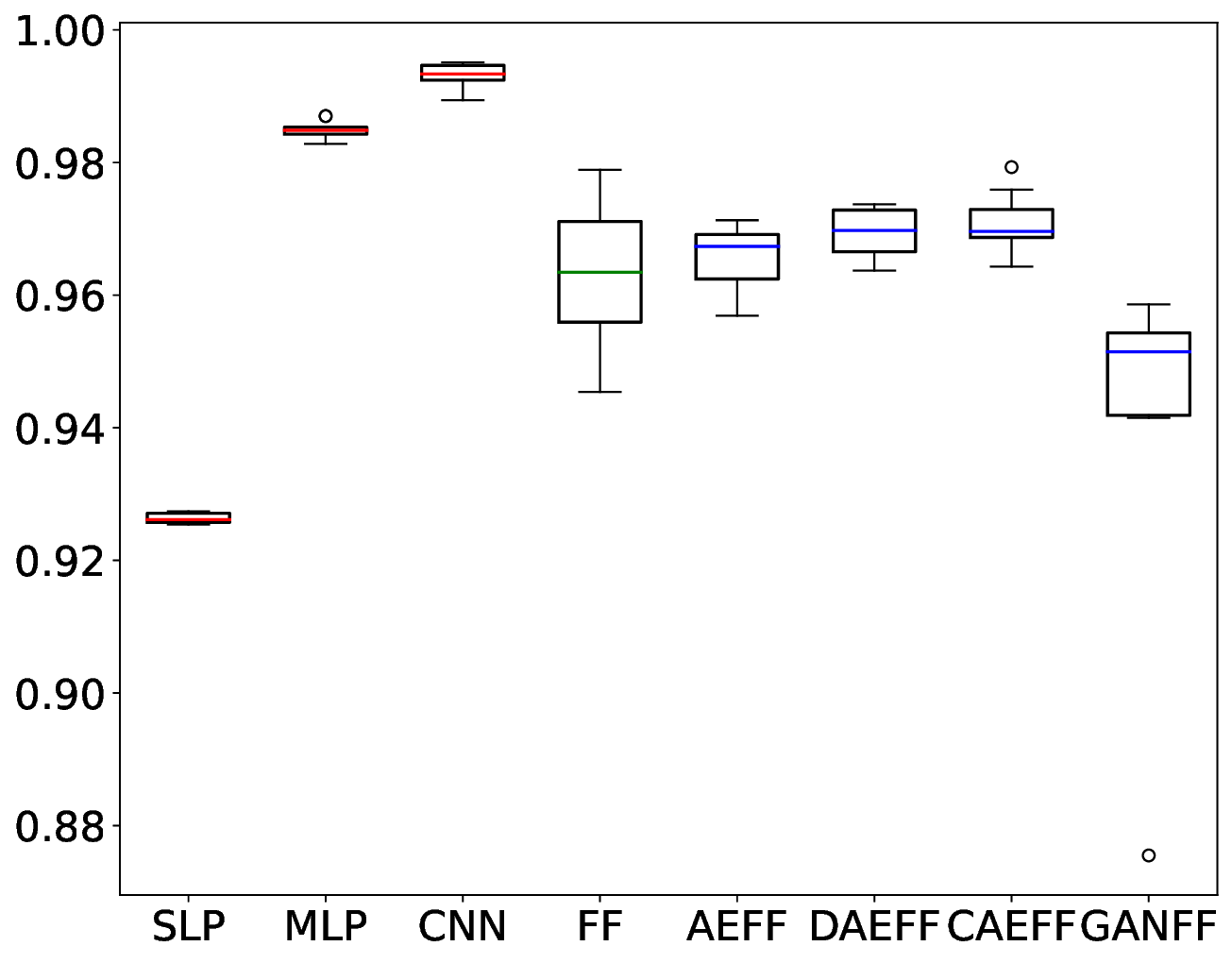}}
\caption{\label{fig:mnist_seq} Performance of sequence training on the MNIST dataset. The $x$-axis represents models and the $y$-axis represents accuracy.\\(From left to right, the $x$-axis displays SLP, MLP, CNN, FF, AEFF, DAEFF, CAEFF, and GANFF.)}
\end{figure*}

\begin{figure*}
\centering
\captionsetup{justification=centering}
\subfloat[\label{fig:mnist2} 2 layers]{\includegraphics[width=0.2\paperwidth]{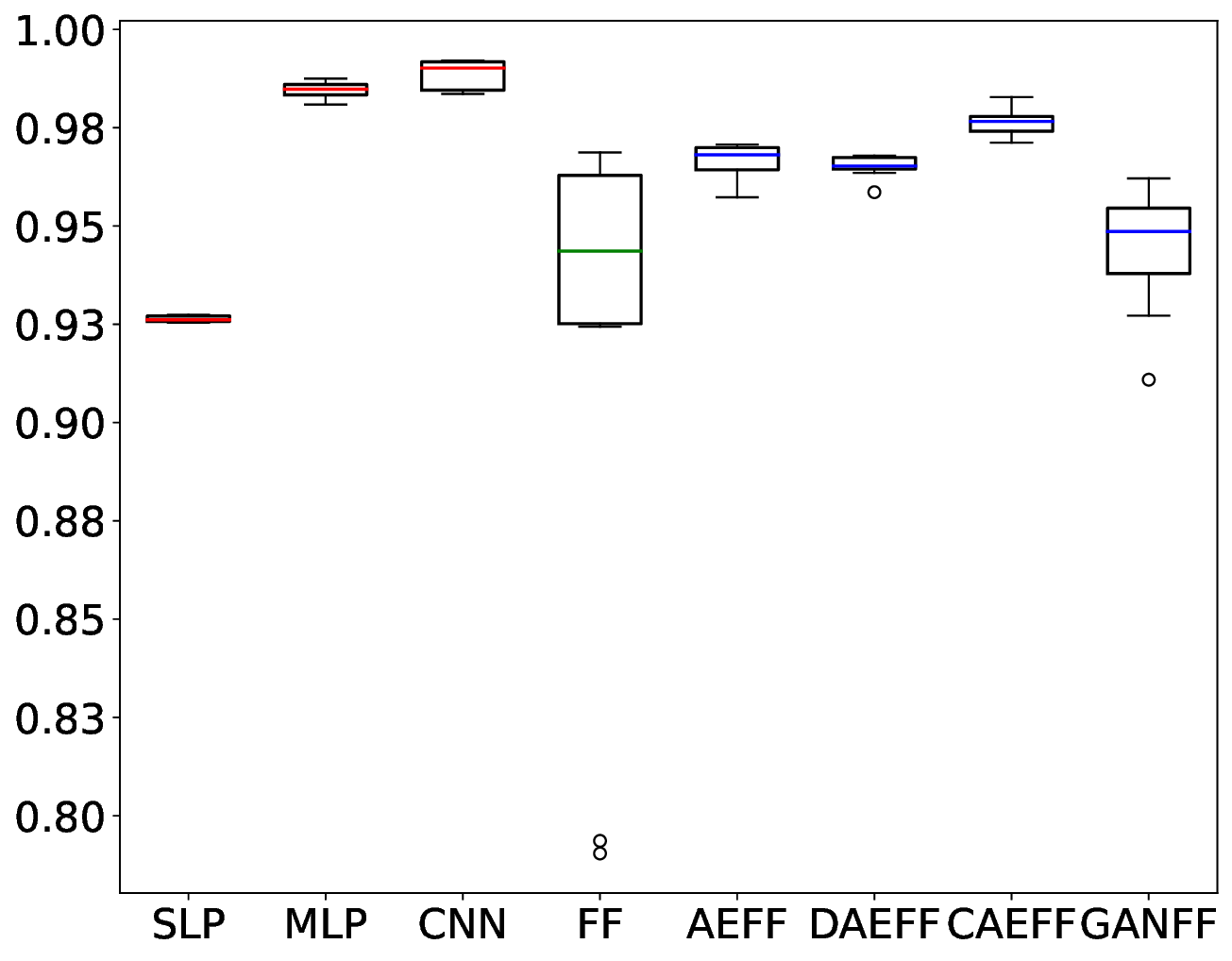}}
\subfloat[\label{fig:mnist3} 3 layers]{\includegraphics[width=0.2\paperwidth]{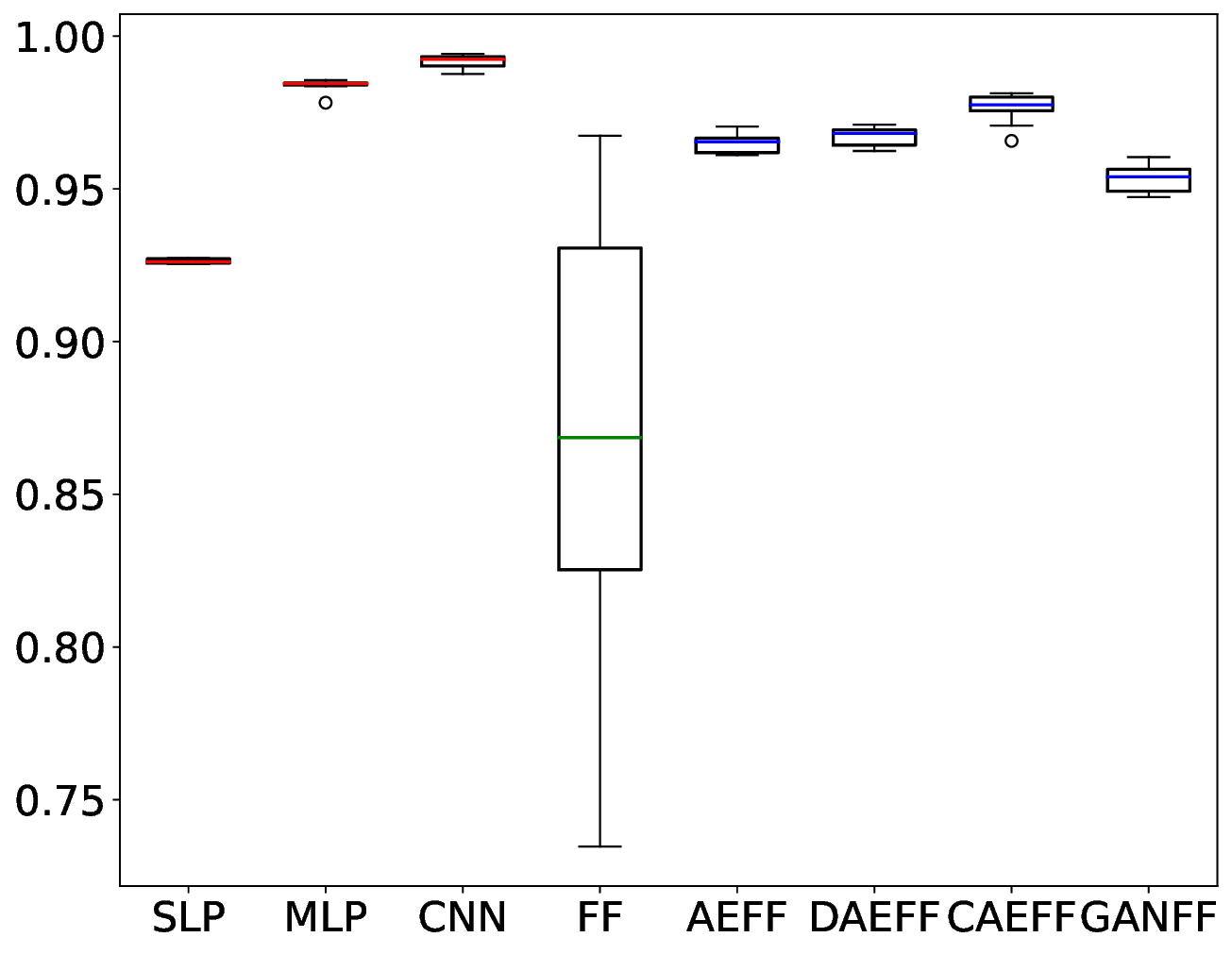}}
\subfloat[\label{fig:mnist4} 4 layers]{\includegraphics[width=0.2\paperwidth]{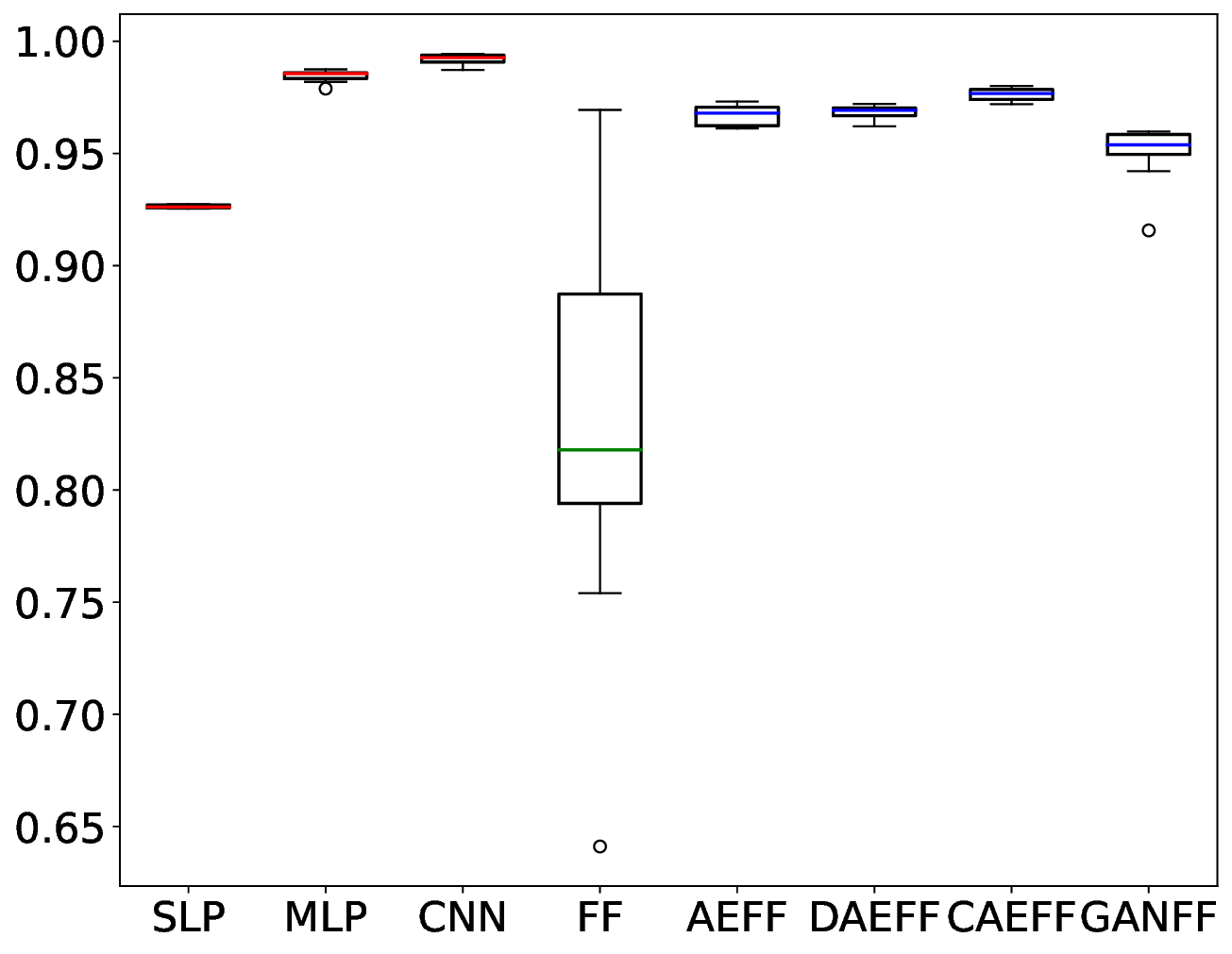}}
\subfloat[\label{fig:mnist5} 5 layers]{\includegraphics[width=0.2\paperwidth]{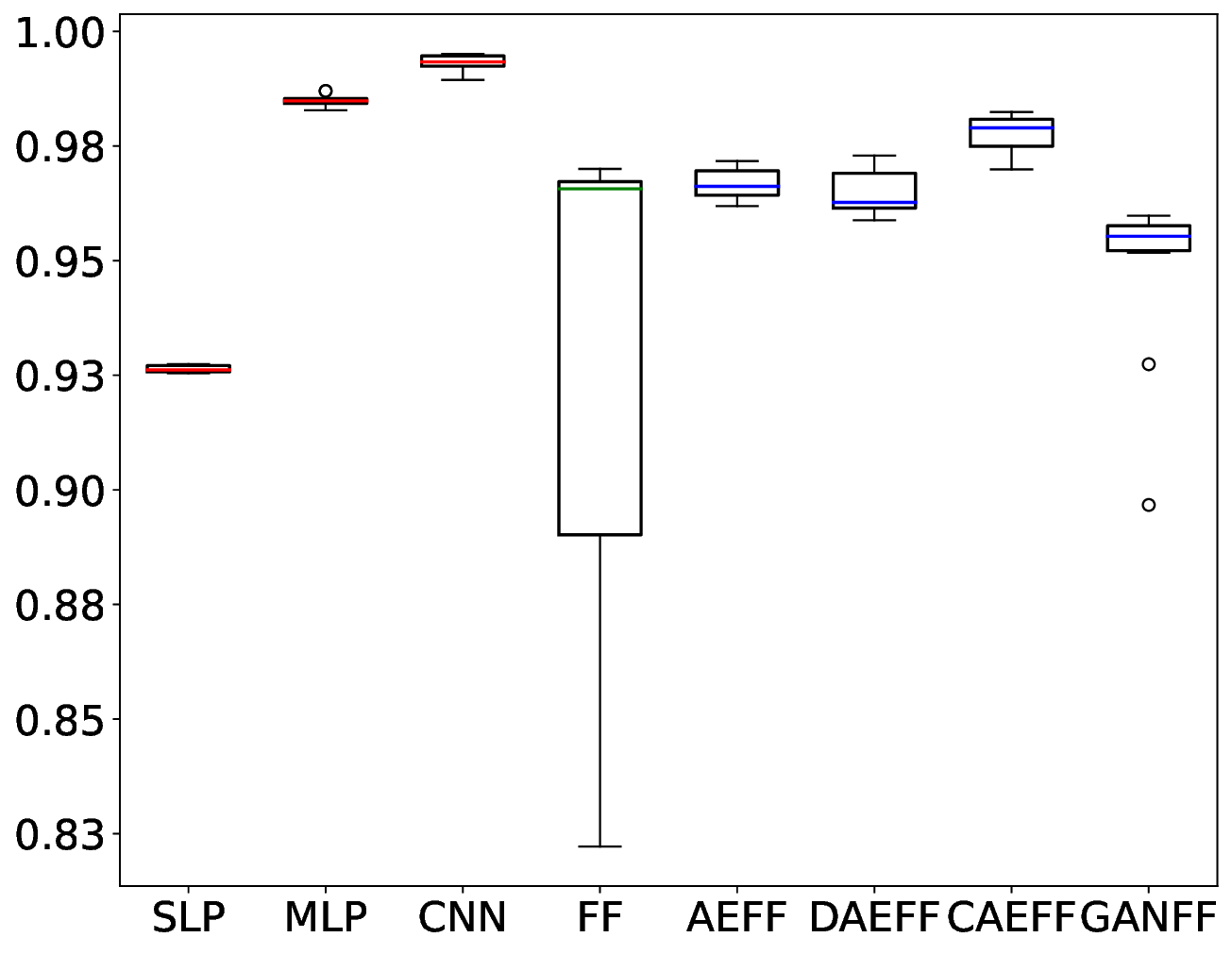}}
\caption{\label{fig:mnist_sep} Performance of separate training on the MNIST dataset. The $x$-axis represents models and the $y$-axis represents accuracy.\\(From left to right, the $x$-axis displays SLP, MLP, CNN, FF, AEFF, DAEFF, CAEFF, and GANFF.))}
\end{figure*}

\begin{figure*}
\centering
\captionsetup{justification=centering}
\subfloat[\label{fig:cifar2} 2 layers]{\includegraphics[width=0.2\paperwidth]{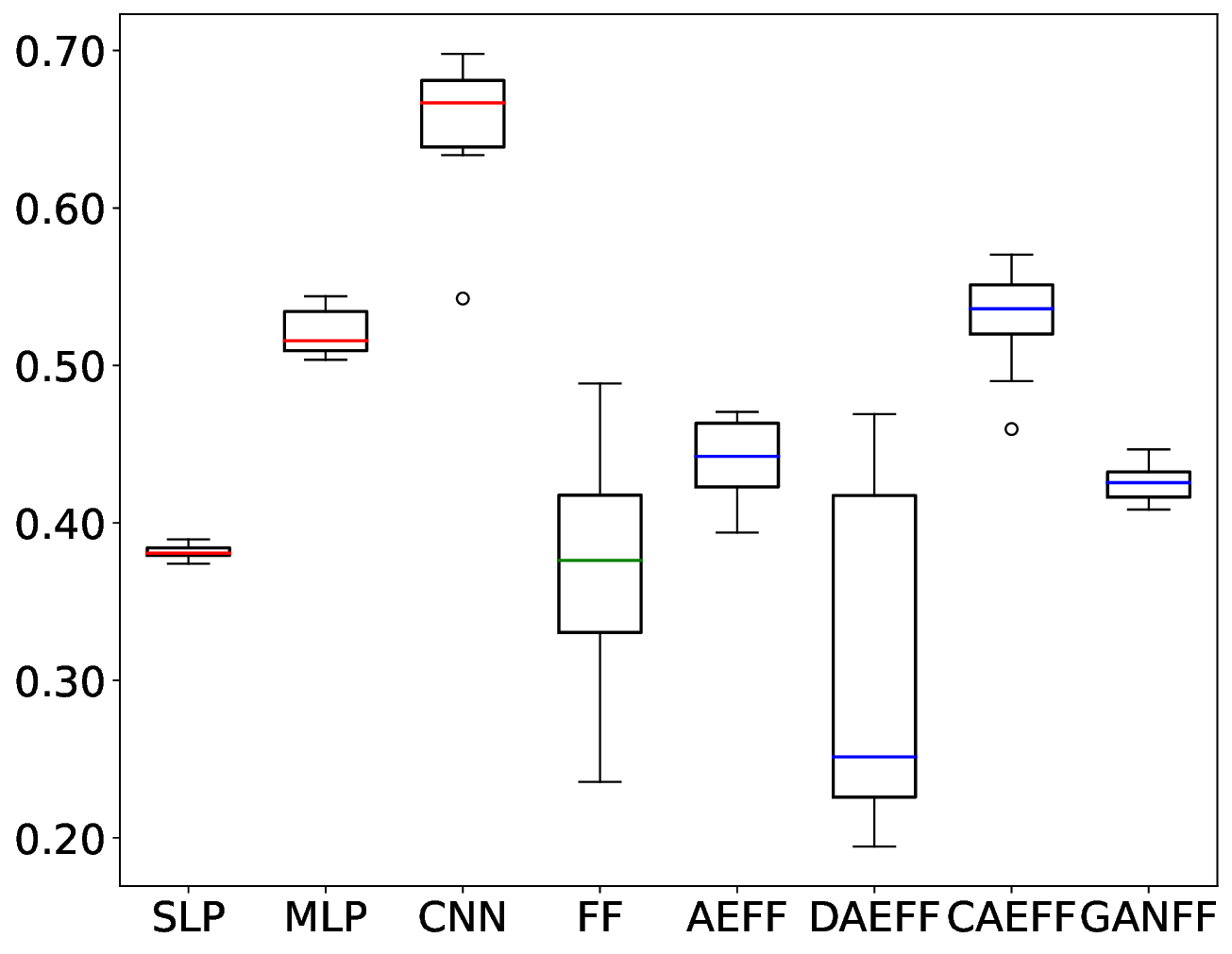}}
\subfloat[\label{fig:cifar3} 3 layers]{\includegraphics[width=0.2\paperwidth]{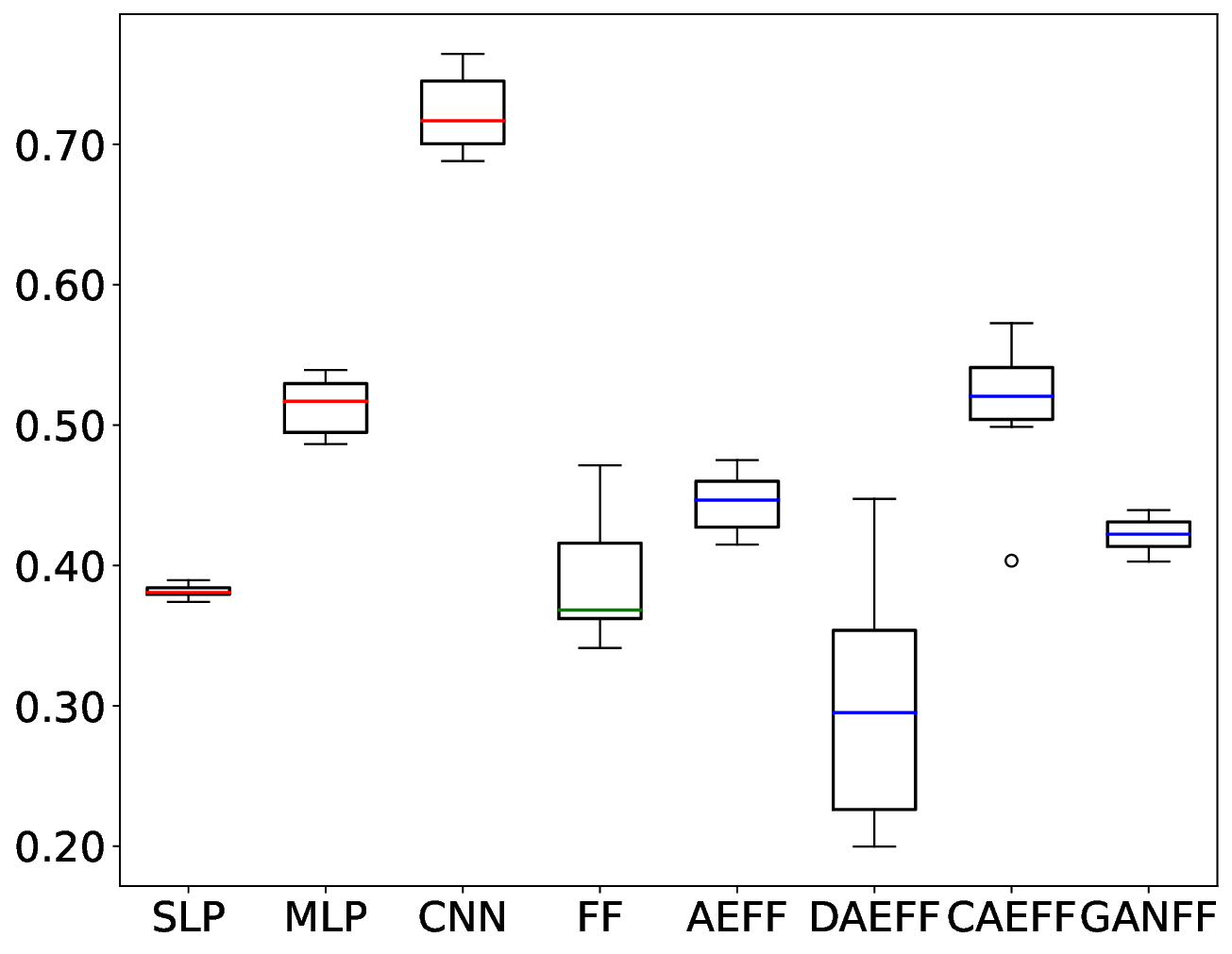}}
\subfloat[\label{fig:cifar4} 4 layers]{\includegraphics[width=0.2\paperwidth]{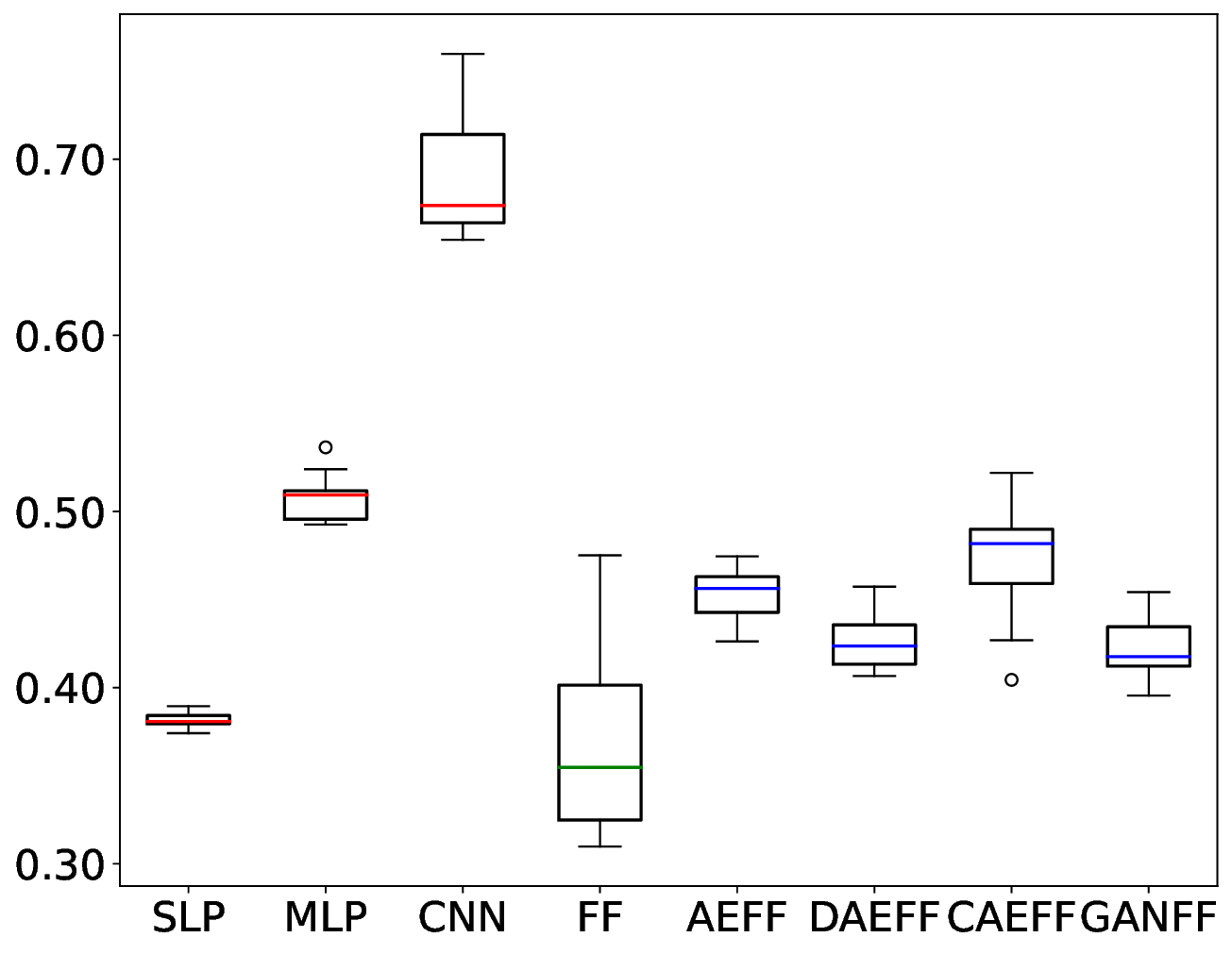}}
\subfloat[\label{fig:cifar5} 5 layers]{\includegraphics[width=0.2\paperwidth]{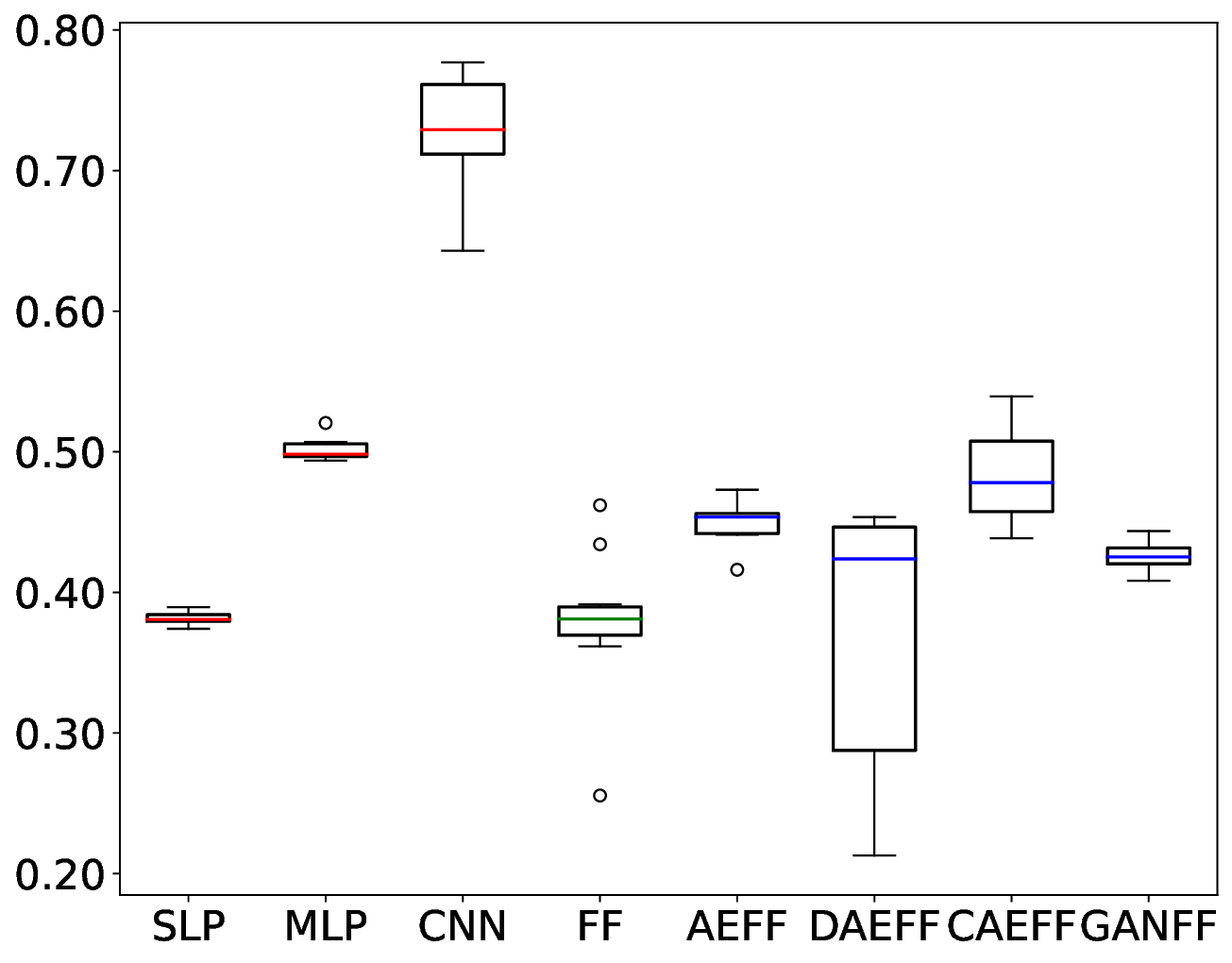}}
\caption{\label{fig:cifar_seq} Performance of sequence training on the CIFAR10 dataset. The $x$-axis represents models and the $y$-axis represents accuracy.\\(From left to right, the $x$-axis displays SLP, MLP, CNN, FF, AEFF, DAEFF, CAEFF, and GANFF.)}
\end{figure*}

\begin{figure*}
\centering
\captionsetup{justification=centering}
\subfloat[\label{fig:cifar2} 2 layers]{\includegraphics[width=0.2\paperwidth]{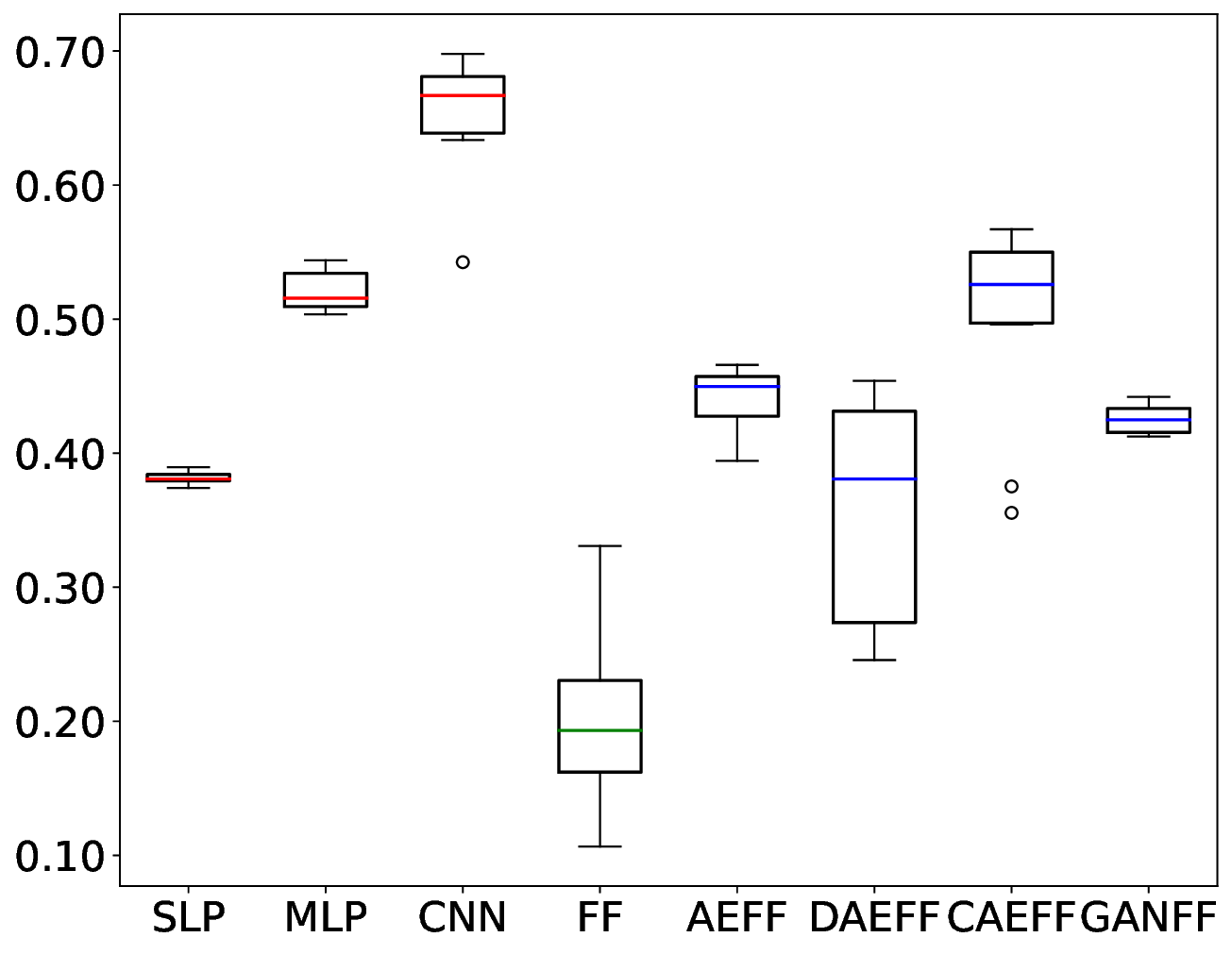}}
\subfloat[\label{fig:cifar3} 3 layers]{\includegraphics[width=0.2\paperwidth]{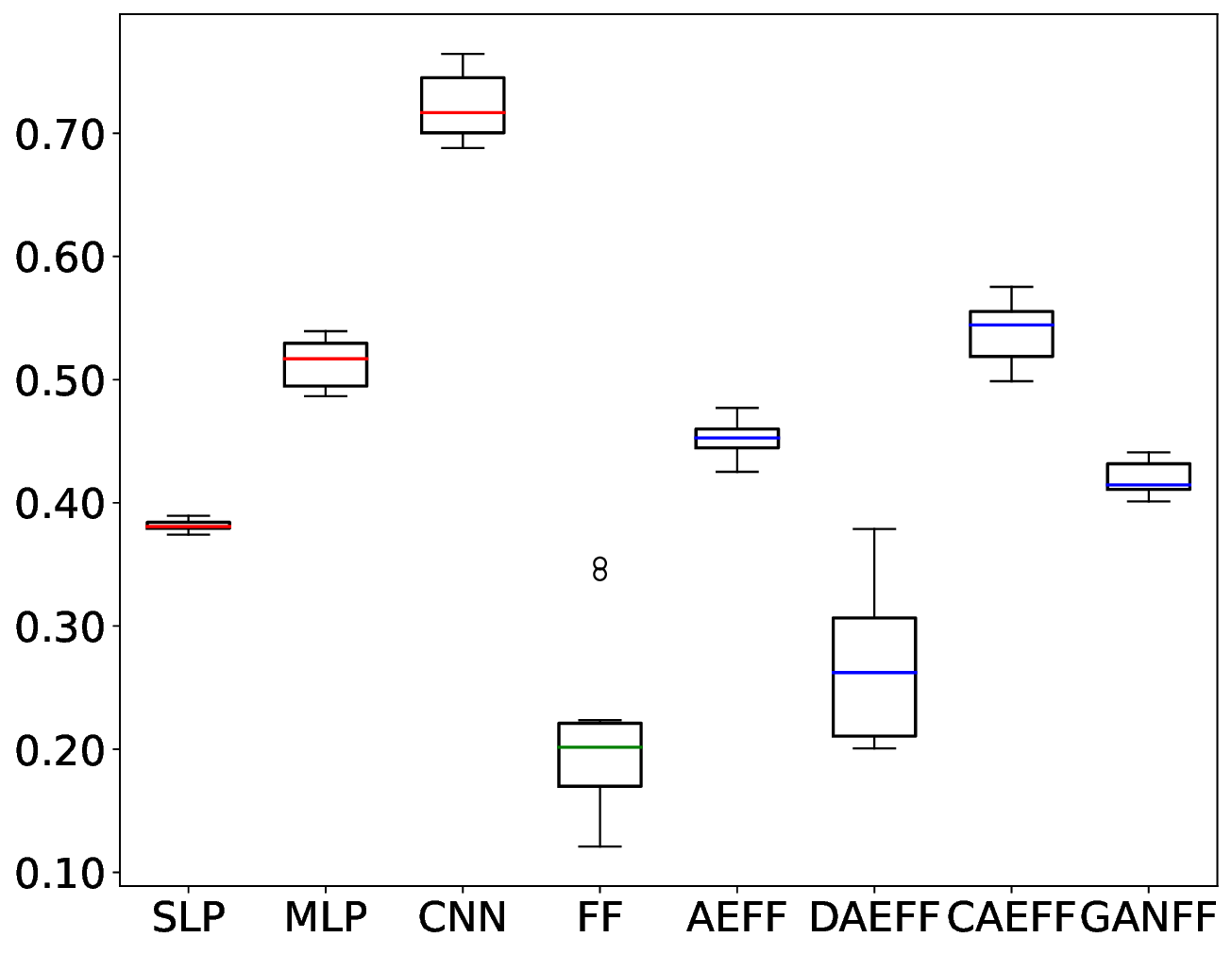}}
\subfloat[\label{fig:cifar4} 4 layers]{\includegraphics[width=0.2\paperwidth]{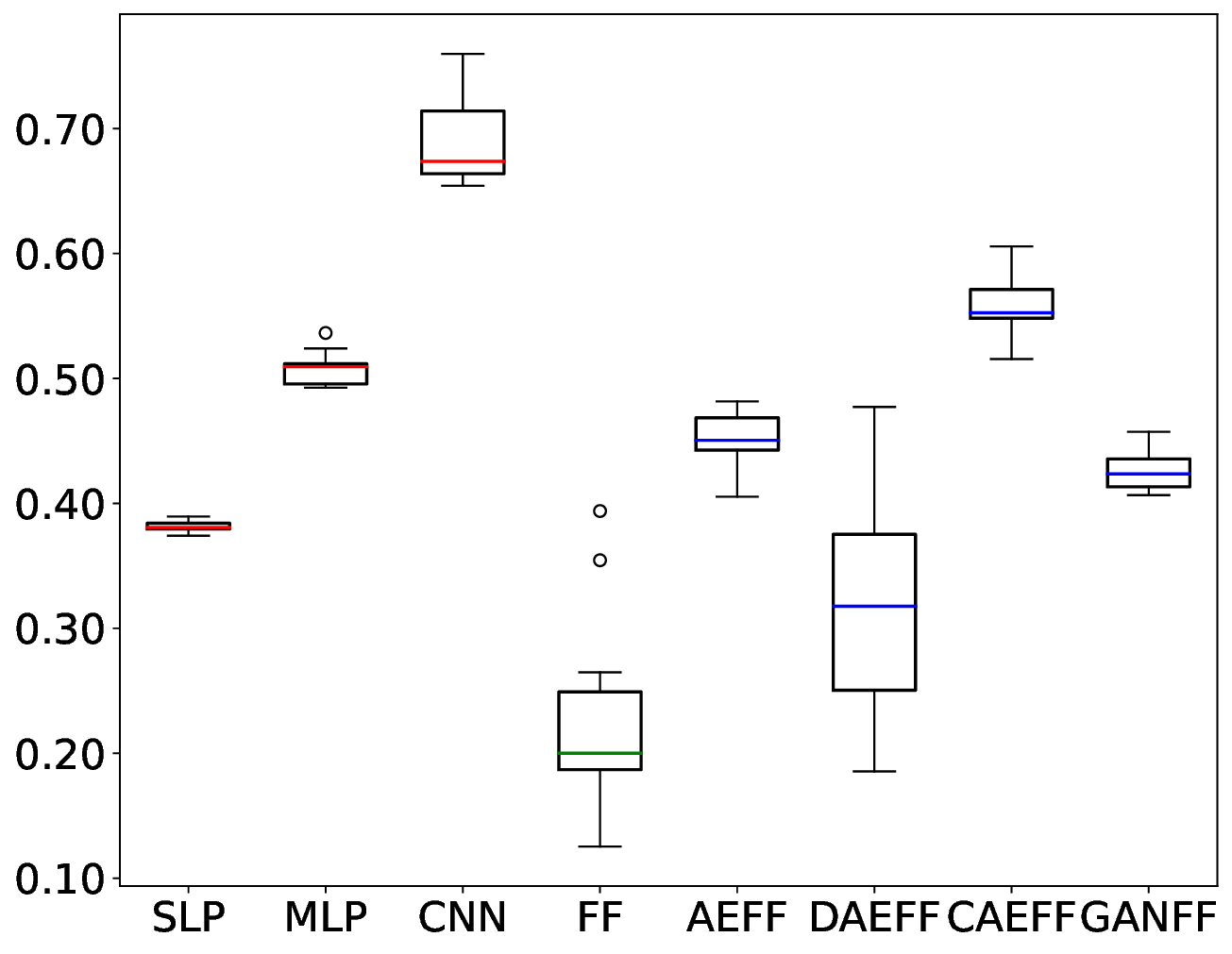}}
\subfloat[\label{fig:cifar5} 5 layers]{\includegraphics[width=0.2\paperwidth]{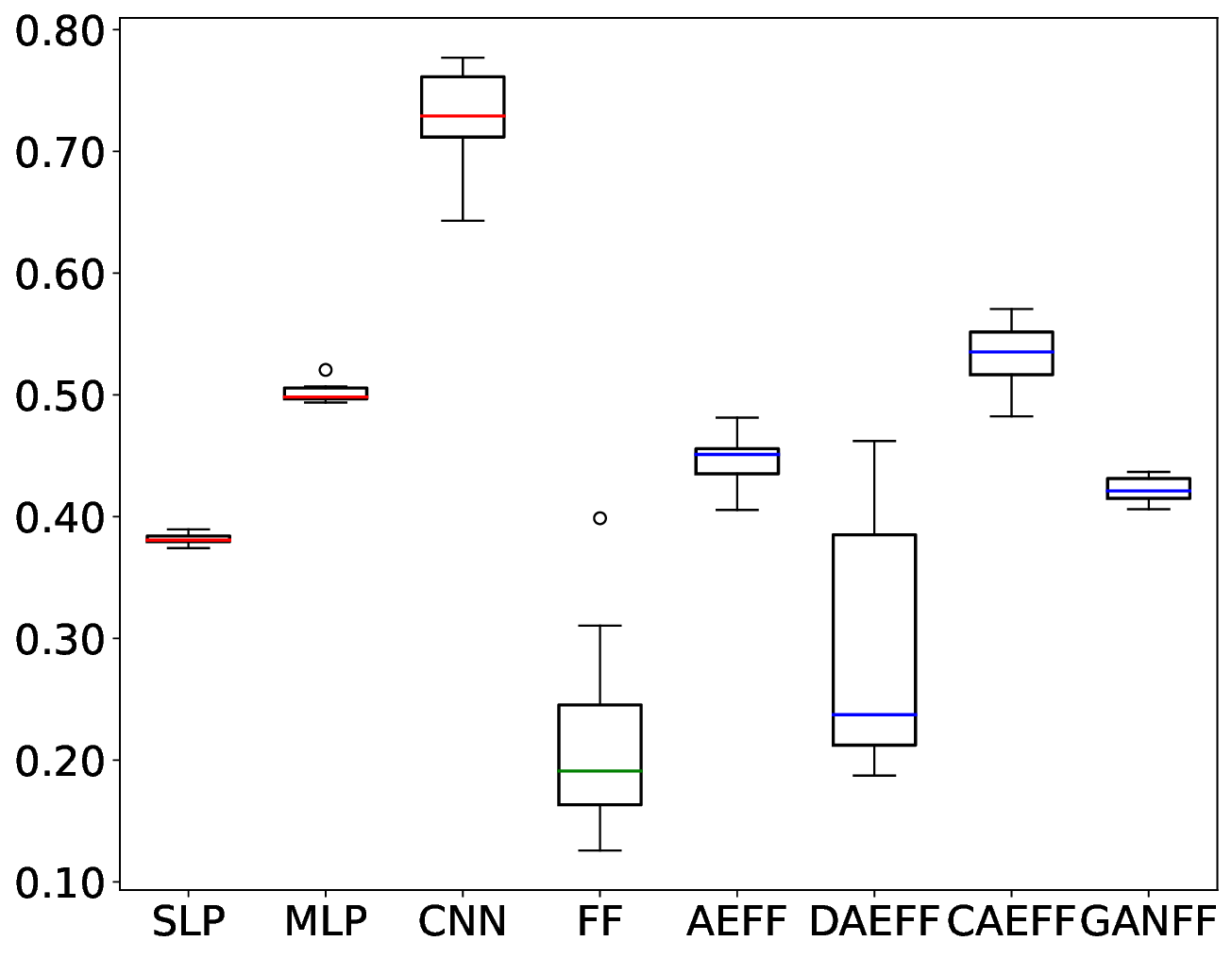}}
\caption{\label{fig:cifar_sep} Performance of separate training on the CIFAR10 dataset. The $x$-axis represents models and the $y$-axis represents accuracy.\\(From left to right, the $x$-axis displays SLP, MLP, CNN, FF, AEFF, DAEFF, CAEFF, and GANFF.)}
\end{figure*}

\subsection{Experimental Results} \label{subsec:ExperimentalResults}

Table \ref{tab:best_mnist} and \ref{tab:best_cifar} present the best performance achieved from 10 measurements conducted through hyper-parameter tuning in all experiments using the sequence and separate training approaches. All performance in this experiment was measured by accuracy. In the overall experiments, CNN trained with BP consistently exhibited the highest performance. On the MNIST dataset, both FF and UFF models showed slightly lower performance than MLP, with CAEFF among them achieving the highest performance. When applied to the CIFAR10 dataset, FF and UFF models generally demonstrated lower performance than MLP; however, CAEFF consistently outperformed MLP.

In the sequence training, where all cells were sequentially trained for one epoch each, FF outperformed UFF models excluding CAEFF. However, in the separate training, where the preceding cells were trained to their maximum epochs before training the next cell, UFF models outperformed sequence training and generally performed better than FF. Notably, FF in the separate training experienced significant performance degradation on the CIFAR10 dataset.

Figure \ref{fig:mnist_seq}, \ref{fig:mnist_sep}, \ref{fig:cifar_seq}, and \ref{fig:cifar_sep} depict box plots of the 10 measurements obtained from hyper-parameter tuning, encompassing the entire range of experimental results. The $y$-axis represents accuracy, and the $x$-axis is labeled sequentially as SLP, MLP, CNN, FF, AEFF, DAEFF, CAEFF, and GANFF. The circular points within the box plots represent outliers, and the colored line inside the box signifies the median value. The range of each box extends from the $25th\ percentile$ to the $75th\ percentile$.

Figure \ref{fig:mnist_seq} represents the performance measured on the MNIST dataset using the sequence training. Except for FF and GANFF, the results generally exhibit a stable distribution. Figure \ref{fig:mnist_sep} displays the performance measured on the MNIST dataset using the separate training. Notably, the performance distribution of FF appears to be quite unstable.

Figure \ref{fig:cifar_seq}, which showcases the performance measured on the CIFAR10 dataset using the sequence training, shows a broader range of performance distribution compared to the MNIST dataset. Notably, the distributions of FF and DAEFF appear to be unstable. Figure \ref{fig:cifar_sep} presents the performance measured on the CIFAR10 dataset using the separate training, and similar to Figure \ref{fig:cifar_seq}, the performance distribution of FF and DAEFF remains unstable. On the other hand, GANFF demonstrates a notably stable distribution in contrast to its performance on the MNIST dataset. The unstable distribution observed in DAEFF and GANFF is likely because these models incorporate random noise during their training.

\begin{table}[]
\begin{adjustbox}{max width=\linewidth}
\begin{tabular}{c|cc|cc}
\hline
               & \multicolumn{2}{c|}{\textbf{MNIST}}                                                                                       & \multicolumn{2}{c}{\textbf{CIFAR10}}                                                                                        \\
               & \begin{tabular}[c]{@{}c@{}}Sequence\\ training\end{tabular} & \begin{tabular}[c]{@{}c@{}}Separate\\ training\end{tabular} & \begin{tabular}[c]{@{}c@{}}Sequence\\ training\end{tabular} & \begin{tabular}[c]{@{}c@{}}Separate\\ training\end{tabular} \\ \hline
\textbf{MLP}   & 0.67                                                        & 0.67                                                        & 0.71                                                        & 0.71                                                        \\
\textbf{CNN}   & 0.66                                                        & 0.66                                                        & 0.67                                                        & 0.67                                                        \\ \hline
\textbf{FF}    & 0.97                                                        & 1.11                                                        & 1.36                                                        & 1.72                                                        \\ \hline
\textbf{AEFF}  & 1.57                                                        & 1.53                                                        & 1.61                                                        & 1.32                                                        \\
\textbf{DAEFF} & 1.51                                                        & 1.45                                                        & 1.62                                                        & 1.78                                                        \\
\textbf{CAEFF} & 2.12                                                        & 2.78                                                        & 2.86                                                        & 4.36                                                        \\
\textbf{GANFF} & 3.83                                                        & 3.91                                                        & 3.78                                                        & 3.74                                                        \\ \hline
\end{tabular}
\end{adjustbox}
\caption{The average training time (in seconds) for 1 epoch across 5 runs for each model with two layers.}
\label{tab:time}
\end{table}

Table \ref{tab:time} represents the training time (in seconds) per epoch measured for models with two layers or cells. The measurements were taken using a single Nvidia V100 GPU, and the presented values are the average of five runs. FF, AEFF and DAEFF take approximately twice as long as MLP and CNN in terms of training time. Also, it can be observed that CAEFF and GANFF require at least 3 to 7 times longer time compared to the others.


\section{Discussion} \label{sec:Discussion}


Through the conducted experiments in this study, we validated the performance and practicality of the proposed UFF models. The UFF models consistently demonstrated superior performance compared to SLP. This finding underscores that even with the sole utilization of the forward pass in unsupervised models, inter-layer information exchange and feedback enable effective learning. However, UFF models generally exhibited lower performance than MLP using BP, indicating the challenge of fully replicating BP's performance solely through the forward pass. Nevertheless, models like CAEFF, which employ convolutions, demonstrated performance surpassing that of MLP, albeit lower than CNN. This implies that our proposed UFF training method, while not outperforming BP under identical conditions and models, has the potential to perform similarly well when using deep learning models that have previously demonstrated success with BP.


Our study verified that the proposed UFF method is capable of training deep learning models via the FF approach, which employs only forward passes similar to biological processes. Additionally, we demonstrated the feasibility of using standard data and a universal loss function with UFF, as opposed to the specialized data and loss calculation methods required by FF. This advancement addresses and overcomes FF's limitation which is difficult to apply to various models, making it possible to use it universally. Experimental results also indicated that while FF exhibits a larger performance deviation compared to BP when the same model is used, our UFF method demonstrates significantly more stable performance than FF in AEFF and CAEFF configurations.


The limitations of the UFF are its lower performance and longer training time in comparison to BP. However, experimental observations in \cite{ff} indicating UFF's superior performance relative to MLP using BP suggest the possibility of UFF attaining performance levels comparable to BP with the identification of optimal hyper-parameter settings. Nonetheless, given the substantial variability in its performance and a training duration approximately twice as long as that of BP, achieving high efficacy with UFF would necessitate additional time and effort on the part of the learner.


Considering the strengths and limitations of the proposed method, it emerges as a feasible alternative to the original FF approach. It effectively addresses the constraints of FF while maintaining stable performance. Additionally, there remains potential for further performance enhancements and applicability to various large-scale models currently demonstrating efficacy, indicating prospects for utilization and universality. Although it cannot completely supplant BP due to its inherent limitations, it offers a viable option in scenarios where BP implementation is challenging. For instance, in federated learning\cite{federatedlearning}, where the physical separation of model layers complicates the use of BP, the UFF method could facilitate more effective sharing of weights and feedback between local and global models.
    
\section{Conclusion} \label{sec:Conclusion}

Our proposed Unsupervised leaerning Forward-Forward methodology utilizing unsupervised learning models demonstrated performance that was comparable to or better than existing Forward-Forward algorithms. Additionally, we confirmed that our model exhibited performance close to that achieved by deep learning models using backpropagation. Moreover, we addressed the issue of low versatility in the traditional Forward-Forward algorithm by solving the problems related to the need for specific forms of input and loss functions. This advancement addressed the problem of unstable performance, which was evident in significant variations within the same model using the Forward-Forward approach. Lastly, our research confirmed the potential applicability of this methodology in areas where conventional deep learning training methods face challenges.

For future research, we plan to apply our proposed training method to a variety of large-scale deep learning models to further verify its versatility. Additionally, we aim to explore the potential applicability under various conditions through experiments in scenarios where backpropagation is challenging to implement, such as in federated learning. Lastly, we intend to enhance the stability of our proposed methodology by integrating it with several alternative approaches that could replace backpropagation, thereby aiming to achieve performance comparable to that of backpropagation.






\bibliographystyle{named}
\bibliography{ijcai24_uff}

\begin{thebibliography}{}

\bibitem[\protect\citeauthoryear{Ba \bgroup \em et al.\egroup }{2016}]{layernorm}
Jimmy~Lei Ba, Jamie~Ryan Kiros, and Geoffrey~E Hinton.
\newblock Layer normalization.
\newblock {\em arXiv preprint arXiv:1607.06450}, 2016.

\bibitem[\protect\citeauthoryear{Biewald and others}{2020}]{wandb}
Lukas Biewald et~al.
\newblock Experiment tracking with weights and biases.
\newblock {\em Software available from wandb. com}, 2:233, 2020.

\bibitem[\protect\citeauthoryear{Crick}{1989}]{bpcritic89}
Francis Crick.
\newblock The recent excitement about neural networks.
\newblock {\em Nature}, 337(6203):129--132, 1989.

\bibitem[\protect\citeauthoryear{Goodfellow \bgroup \em et al.\egroup }{2014}]{gan}
Ian Goodfellow, Jean Pouget-Abadie, Mehdi Mirza, Bing Xu, David Warde-Farley, Sherjil Ozair, Aaron Courville, and Yoshua Bengio.
\newblock Generative adversarial nets.
\newblock {\em Advances in neural information processing systems}, 27, 2014.

\bibitem[\protect\citeauthoryear{Grossberg}{1987}]{bpcritic87}
Stephen Grossberg.
\newblock Competitive learning: From interactive activation to adaptive resonance.
\newblock {\em Cognitive science}, 11(1):23--63, 1987.

\bibitem[\protect\citeauthoryear{Hinton and Zemel}{1993}]{autoencoder}
Geoffrey~E Hinton and Richard Zemel.
\newblock Autoencoders, minimum description length and helmholtz free energy.
\newblock {\em Advances in neural information processing systems}, 6, 1993.

\bibitem[\protect\citeauthoryear{Hinton}{2022}]{ff}
Geoffrey Hinton.
\newblock The forward-forward algorithm: Some preliminary investigations.
\newblock {\em arXiv preprint arXiv:2212.13345}, 2022.

\bibitem[\protect\citeauthoryear{Hubel and Wiesel}{1962}]{rc62}
David~H Hubel and Torsten~N Wiesel.
\newblock Receptive fields, binocular interaction and functional architecture in the cat's visual cortex.
\newblock {\em The Journal of physiology}, 160(1):106, 1962.

\bibitem[\protect\citeauthoryear{Kelley}{1960}]{bp60}
Henry~J Kelley.
\newblock Gradient theory of optimal flight paths.
\newblock {\em Ars Journal}, 30(10):947--954, 1960.

\bibitem[\protect\citeauthoryear{Konečný \bgroup \em et al.\egroup }{2016}]{federatedlearning}
Jakub Konečný, H.~Brendan McMahan, Felix~X. Yu, Peter Richtarik, Ananda~Theertha Suresh, and Dave Bacon.
\newblock Federated learning: Strategies for improving communication efficiency.
\newblock In {\em NIPS Workshop on Private Multi-Party Machine Learning}, 2016.

\bibitem[\protect\citeauthoryear{LeCun \bgroup \em et al.\egroup }{1989}]{cnn}
Yann LeCun, Bernhard Boser, John~S Denker, Donnie Henderson, Richard~E Howard, Wayne Hubbard, and Lawrence~D Jackel.
\newblock Backpropagation applied to handwritten zip code recognition.
\newblock {\em Neural computation}, 1(4):541--551, 1989.

\bibitem[\protect\citeauthoryear{Lillicrap \bgroup \em et al.\egroup }{2014}]{fa}
Timothy~P Lillicrap, Daniel Cownden, Douglas~B Tweed, and Colin~J Akerman.
\newblock Random feedback weights support learning in deep neural networks.
\newblock {\em arXiv preprint arXiv:1411.0247}, 2014.

\bibitem[\protect\citeauthoryear{Linnainmaa}{1970}]{bp70}
Seppo Linnainmaa.
\newblock {\em The representation of the cumulative rounding error of an algorithm as a Taylor expansion of the local rounding errors}.
\newblock PhD thesis, Master’s Thesis (in Finnish), Univ. Helsinki, 1970.

\bibitem[\protect\citeauthoryear{Loshchilov and Hutter}{2019}]{adamw}
Ilya Loshchilov and Frank Hutter.
\newblock Decoupled weight decay regularization.
\newblock In {\em International Conference on Learning Representations}, 2019.

\bibitem[\protect\citeauthoryear{Marblestone \bgroup \em et al.\egroup }{2016}]{bpcritic16}
Adam~H Marblestone, Greg Wayne, and Konrad~P Kording.
\newblock Toward an integration of deep learning and neuroscience.
\newblock {\em Frontiers in computational neuroscience}, 10:94, 2016.

\bibitem[\protect\citeauthoryear{Millidge \bgroup \em et al.\egroup }{2022}]{pc22}
Beren Millidge, Alexander Tschantz, and Christopher~L Buckley.
\newblock Predictive coding approximates backprop along arbitrary computation graphs.
\newblock {\em Neural Computation}, 34(6):1329--1368, 2022.

\bibitem[\protect\citeauthoryear{Nair and Hinton}{2010}]{relu}
Vinod Nair and Geoffrey~E Hinton.
\newblock Rectified linear units improve restricted boltzmann machines.
\newblock In {\em Proceedings of the 27th international conference on machine learning (ICML-10)}, pages 807--814, 2010.

\bibitem[\protect\citeauthoryear{Ororbia and Mali}{2023}]{pff}
Alexander Ororbia and Ankur~A Mali.
\newblock The predictive forward-forward algorithm.
\newblock In {\em Proceedings of the Annual Meeting of the Cognitive Science Society}, volume~45, 2023.

\bibitem[\protect\citeauthoryear{Rao and Ballard}{1999}]{pc}
Rajesh~PN Rao and Dana~H Ballard.
\newblock Predictive coding in the visual cortex: a functional interpretation of some extra-classical receptive-field effects.
\newblock {\em Nature neuroscience}, 2(1):79--87, 1999.

\bibitem[\protect\citeauthoryear{Rumelhart \bgroup \em et al.\egroup }{1985}]{rnn}
David~E Rumelhart, Geoffrey~E Hinton, Ronald~J Williams, et~al.
\newblock Learning internal representations by error propagation, 1985.

\bibitem[\protect\citeauthoryear{Rumelhart \bgroup \em et al.\egroup }{1988}]{mlp}
David~E Rumelhart, James~L McClelland, PDP~Research Group, et~al.
\newblock Parallel distributed processing.
\newblock {\em Foundations}, 1, 1988.

\bibitem[\protect\citeauthoryear{Shepherd}{1990}]{bpcritic90}
Gordon~M Shepherd.
\newblock The significance of real neuron architectures for neural network simulations.
\newblock {\em Computational neuroscience}, pages 82--96, 1990.

\bibitem[\protect\citeauthoryear{Suzuki and Amaral}{1994}]{rc94}
Wendy~A Suzuki and David~G Amaral.
\newblock Topographic organization of the reciprocal connections between the monkey entorhinal cortex and the perirhinal and parahippocampal cortices.
\newblock {\em Journal of Neuroscience}, 14(3):1856--1877, 1994.

\bibitem[\protect\citeauthoryear{Vincent \bgroup \em et al.\egroup }{2010}]{dae}
Pascal Vincent, Hugo Larochelle, Isabelle Lajoie, Yoshua Bengio, Pierre-Antoine Manzagol, and L{\'e}on Bottou.
\newblock Stacked denoising autoencoders: Learning useful representations in a deep network with a local denoising criterion.
\newblock {\em Journal of machine learning research}, 11(12), 2010.

\bibitem[\protect\citeauthoryear{Whittington and Bogacz}{2017}]{pc17}
James~CR Whittington and Rafal Bogacz.
\newblock An approximation of the error backpropagation algorithm in a predictive coding network with local hebbian synaptic plasticity.
\newblock {\em Neural computation}, 29(5):1229--1262, 2017.

\end{thebibliography}

\end{document}